\documentclass[11pt,a4paper]{article}
\usepackage[hyperref]{emnlp2020}
\usepackage{times}
\usepackage{latexsym}
\usepackage{url}
\usepackage[T1]{fontenc}
\usepackage{graphicx}
\usepackage{amssymb}
\usepackage{epsfig,endnotes,graphicx,subcaption,xspace}
\usepackage{wrapfig}
\usepackage{microtype}
\usepackage{multirow}
\usepackage{multicol}
\usepackage{enumitem}

\usepackage{microtype}

\aclfinalcopy 
\def\aclpaperid{2823} 

\title{Modeling Preconditions in Text with a Crowd-sourced Dataset}

\author{
 Heeyoung Kwon$^1$, Mahnaz Koupaee$^1$, \\
 \textbf{Pratyush Singh$^1$, Gargi Sawhney$^1$, Anmol Shukla$^1$, Keerthi Kumar Kallur$^1$, }\\
 \textbf{Nathanael Chambers$^2$ and Niranjan Balasubramanian$^1$}\\
 \\
 $^1$ Stony Brook University, Stony Brook, New York \\
 $^2$ US Naval Academy, Annapolis, MD \\
 \texttt{\{heekwon, mkoupaee, niranjan\}@cs.stonybrook.edu} \\
 \texttt{nchamber@usna.edu} \\
}

\date{}
\newcommand{\eat}[1]{}
\newcommand{\peko}{{PeKo\xspace}}

\newcommand{\hk}[1]{\textcolor{magenta}{$_{HK}$[#1]}}

\definecolor{navyblue}{rgb}{0.2, 0.2, 0.6}

\newcommand{\nblue}[1]{\textcolor{navyblue}{\textbf{#1}}}

\definecolor{RoseQuartzBg}{HTML}{F7CAC9}
\definecolor{RoseQuartz}{HTML}{F5A798}
\definecolor{Serenity}{HTML}{92A8D1}
\definecolor{OrangeRed}{rgb}{1.0, 0.27, 0.0}
\definecolor{Turquoise}{HTML}{0F4C81}
\usepackage{xparse}
\NewDocumentCommand{\nate}{ mO{} }{\textcolor{OrangeRed}{\textsuperscript{\textit{Nate}}\textsf{\textbf{\small[#1]}}}}

\usepackage{array}

\begin{document}
\maketitle
\begin{abstract}
Preconditions provide a form of logical connection between events that explains {why} some events occur together and information that is complementary to the more widely studied relations such as causation, temporal ordering, entailment, and discourse relations.
Modeling preconditions in text has been hampered in part due to the lack of large scale labeled data grounded in text. This paper introduces \peko, a crowd-sourced annotation of \emph{preconditions} between event pairs in newswire, an order of magnitude larger than prior text annotations. To complement this new corpus, we also introduce two challenge tasks aimed at modeling preconditions: (i) Precondition Identification -- a standard classification task defined over pairs of event mentions, and (ii) Precondition Generation -- a generative task aimed at testing a more general ability to reason about a given event. Evaluation on both tasks shows that modeling preconditions is challenging even for today's large language models (LM). This suggests that precondition knowledge is not easily accessible in LM-derived representations alone. Our generation results show that fine-tuning an LM on \peko\ yields better conditional relations than when trained on raw text or temporally-ordered corpora.

\end{abstract}
\section{Introduction}
Recognizing logical connections between events in text is important for comprehensive document understanding and to improve global coherence in language generation systems. There is a rich body of work in identifying relations between textual events which covers causation \cite{mirza:2014}, temporal relations \cite{timebank}, textual entailment \cite{dagan2005pascal}, and discourse relations \cite{blair:2006}. 

In this work, we focus on the precondition relation, which offers a general view of \emph{why} certain events occur together in the world. This is not easily deduced from other event-event relations. Temporal ordering systems can sequence the order in which events occurred \cite{bethard2013cleartk,chambers2014dense,han:2019} but can't explain why they occurred at all. Which events in a sequence were by chance, and which were required? Textual entailment identifies event paraphrases \cite{berant2015efficient} and some causation \cite{girju:2003}, but their view misses the broader look at enabling events like preconditions. Let the following serve as an example:

\begin{quote}
    I heard a bird \textbf{sing} above as I \textbf{turned} the key in the door. It \textbf{opened} with a \textbf{push}.
\end{quote}

You can sequence these four events in order, but an ordering does not understand the \emph{why} of the situation. One of these events (sing) is clearly not relevant to the door opening. How do we know that \emph{turning} the key is a precondition to \emph{opened} and not \emph{push}? Turning the key usually doesn't cause the door to open (perhaps on some doors, but here a \emph{push} was needed). Turning is simply a precondition. Causation and entailment do not apply to \emph{turn} either. Preconditions thus provide a unique and still fine-grained understanding of this situation. 

How do we build models that can recognize (and learn from) this type of common-sense knowledge in text? Do language models trained on vast amounts of data already capture it? Since there are no large scale datasets that can effectively answer these questions, we introduce \textbf{\peko}, the \textbf{P}r\textbf{e}condition \textbf{K}n\textbf{o}wledge dataset. We also introduce two tasks -- one aimed at recognizing preconditions in text, and the other at generating precondition events for any given target event. 


The core contribution in this paper is this new publicly available crowd-sourced \peko\ dataset. It consists of 28,948 event pairs annotated with precondition relations. 
We will first present our working definition of preconditions, and then discuss how to practically get crowd workers to identify them in text.
We provide analysis of the new corpus and compare it against other existing corpora.

In addition to the corpus, this paper proposes two new challenge tasks.
The first is a traditional classification task on the corpus itself. We thus address critical questions of how to model precondition knowledge. For instance, do today's large language models (e.g., BERT or XLNet) already capture precondition knowledge, and how do they perform on a precondition prediction task? Second, does textual context assist in precondition prediction?
We experiment with varying levels of context and show that identifying preconditions requires careful modeling of the context.

The second proposed task is a precondition \emph{generation} evaluation: models must generate necessary preconditions for a given target event. 
This is a test for how well models can reason about the necessary preconditions for a given situation, which is a useful capability for story generation and learning generalized scripts. We show how \peko\ can be used to train (fine-tune) standard generative models, such as GPT-2, for this task. Empirical results show that fine-tuning on the \peko-derived training set generates at least twice as many preconditions as compared to training on general instances.

All code and data are available at \url{https://stonybrooknlp.github.io/PeKo/}.



\section{Related Work} \label{sec:related}

There has been a vast amount of research on extracting different types of relations between events including temporal~\cite{timebank}, causal \cite{girju2003automatic}, and paraphrasal relationships \cite{lin2001dirt}, but relatively less research into precondition relationships. One of the early definitions and computational use of preconditions comes from the STRIPS program \cite{fikes1971strips}. Preconditions were defined as a set of conditions that MUST be met in order for the action (event) to be allowed to take place. 

Later work focused on aggregating precondition knowledge for a small class of action words, leveraging FrameNet and a text corpus to generate candidate precondition words using a PMI-based heuristic~\cite{sil2010extracting,sil2011extracting}. Using small amounts of labeled data, they use hand-crafted PMI and wordnet based features to learn a SVM-based classifier that scores preconditions for a given action.  
Branavan et al.\ \shortcite{branavan2012learning} learned domain-specific preconditions from written instructions for the game of Minecraft. 
The instructions are procedural and well suited for identification. These mostly target preconditions that are event-state relations as opposed to our goals of textual event-event identification.

ATOMIC~\cite{sap2019atomic} is a related crowd-sourced dataset of event-event relations, where given a simple target event (verb phrase and its arguments), crowd workers provided various types of common-sense knowledge. This included `NEED' events analogous to our precondition events for a target. The main difference is our work grounds both target and precondition events in news text, whereas ATOMIC elicits general world knowledge, a complementary approach with different trade-offs. Interestingly, we find that the precondition relations learnt from textually grounded news events generalize to story events in ATOMIC for our generation task.


\noindent
\textbf{Annotated Text Corpora} Three existing datasets capture some form of precondition knowledge in their annotations: the Rich Event Description (RED) dataset \cite{o2016richer}, CaTeRS \cite{mostafazadeh2016caters}, and Event StoryLine \cite{caselli2017event}. 
These are generally too small for learning text classifiers as we briefly describe now.

RED is the most directly related, created to model a broad set of event-event relations in news. Preconditions are not their sole focus, though, so this dataset only contains \textasciitilde1000 precondition instances. CaTeRs shares a similar problem to RED. It has an \texttt{enables} relation similar to precondition, but since the domain is 5-sentence short stories and preconditions aren't the main focus, it only has \textasciitilde400 instances. The Event StoryLine dataset is small in size too, but also doesn't have a precise precondition relation. The dataset instead has RISING\_ACTION that includes preconditions in its definition, but the same label captures other concepts like subevents and entailment. There are \textasciitilde5000 instances, but only a fraction are preconditions and it is not possible to separate them out.

This paper is thus unique to prior work by annotating grounded written text at a scale large enough to enable machine learning solutions.
This enables our target tasks: text classification and generation.



%

%

\section{Preconditions as Relations}
Our goal is to develop a resource that can help models reason about the necessary preconditions for events mentioned in text. This is useful for planning towards a goal, explaining how a certain situation came about, and predicting what future events are plausible.  We make two important design choices in building such a resource: \textbf{Grounding} -- the resource is grounded in text, particularly over events in the news domain, and {\bf Framing} -- we construct the resource with preconditions framed as event-to-event relation pairs in a specific context.\\\vspace{-1mm}

\noindent{\bf Grounding:} We ground the resource to text so that we can leverage the full context of the events, and we choose the news domain due to its common use in other event-related tasks such as event extraction, schema generation, and temporal reasoning.\\\vspace{-1mm}

\noindent{\bf Framing:} Broadly speaking, preconditions specify what must exist/happen before something else can exist/happen \cite{fikes1971strips,sap2019atomic}. It is natural to think of a precondition as a state of the world that must be satisfied for an event to happen i.e. a state-event relation. However, the state of the world is hard to circumscribe for most real world events, and more importantly the precondition state is often left unsaid in a story. Rather, the author will more often mention an event from which it follows that the precondition state is satisfied. Thus, it makes sense to frame preconditions as relations between \emph{two events} described in their specific textual context.

We first present a formal definition based on this notion and then describe a crowdsourcing methodology for obtaining this knowledge at scale.\\\vspace{-1mm}

\noindent{\bf Definition:} Given a target event mention $t$ and a candidate event mention $p$, we assert $p$ \emph{is a precondition event for} $t$ if $p$ is necessary for $t$ to happen i.e., $t$ likely would not have occurred without $p$, in the current text context.\\\vspace{-1mm}

Using the example of opening a door from the Introduction, turning the key is a precondition event (for opening the door) because it results in a state where the door is unlocked. The opening event cannot occur without such a state. Importantly, we do not define a precondition event as an absolute requirement for the target (the door opening) to occur in all scenarios. 
However, we do require that the target event \emph{likely} would \emph{not} have occurred in the \emph{current context}. 
This allows another story with an alternate event, such as ``I picked the lock''. Both picking-lock and turning-key are preconditions in their own story contexts.
Strict logicians might take issue, but language understanding requires a looser definition that uses likelihood of occurrence when interpreting real-world scenarios.

\section{Preconditions Dataset}

This section describes our methodology to annotate news articles with the previous section's definitions.
One problem with annotating preconditions in text is the large number of event mentions in each article, which means annotation of all possible event pairs is infeasible. The temporal community has struggled with this same dilemma \cite{chambers2014dense,vash:2019}.

We address the question of which pairs to annotate with two approaches. First, instead of attempting a dense annotation of \emph{few} articles, we sub-sample candidate pairs of events across \emph{many} articles. Second, we use an automatic temporal relation classifier to filter pairs by identifying possible candidates. We then ask crowd-workers to annotate the resulting pairs for preconditions.

\subsection{Candidate Event Pair Extraction}

Sub-sampling event pairs at random from a document can result in a large number of pairs that are not preconditions. Because precondition event pairs ought to be temporally related (i.e., the precondition should precede the target event), we can filter the candidate event pairs to only those that are in a BEFORE or AFTER relationship.

As a first step, we extract events and their temporal relations from news articles using~CAEVO~\cite{chambers2014dense}, a temporal relation extraction system. We chose CAEVO over other available systems for two main reasons, although it's not the only option out there: (1) it automatically extracts \emph{both} events and their temporal relations, and (2) it extracts events in any form (verbs, nouns, and adjectives), which gives a broader coverage than some other recent systems that only consider verbs as events~\cite{ning2018improving}. We used CAEVO on a random sample of 6,837 articles in the New York Times Annotated Corpus~\cite{nytcorpus}. 

On average CAEVO extracted around 63 events per article, which yielded a total of 3,906 possible relation candidates per document. We filtered these to retain only pairs of events that have a BEFORE or AFTER temporal relation between them. We call the temporally preceding event the {\em candidate precondition}, and the temporally subsequent event in the pair the {\em target event}. We filtered out pairs involving causative targets or reporting verb preconditions to remove trivial context independent preconditions (see Appendix for examples).

From the remaining, we randomly sampled 40,500 pairs for annotation. We used the first 500 in a pilot annotation to help us improve the task instructions. We then used the remainder for the actual annotation.

\begin{figure}[t!]
    \centering
    \includegraphics[width=0.48\textwidth]{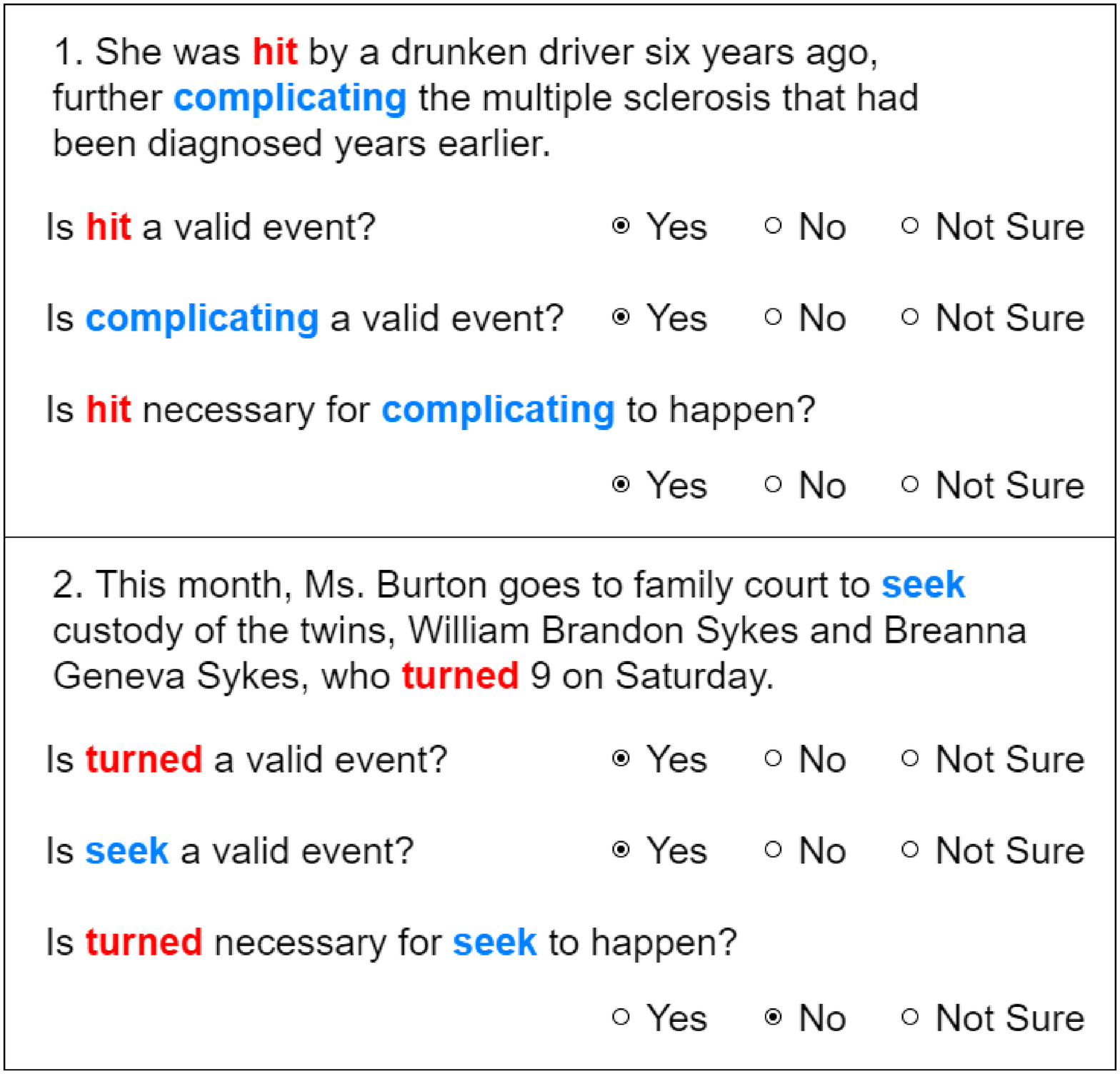}
    \vspace{-.25in}
    \caption{Example instances annotated by crowd-workers. Each HIT included ten such instances.}
    \label{fig:mturk_example}
    \vspace{-.2in}
\end{figure}

\subsection{Crowdsourcing}\label{sec:crowdsourcing}
The annotators were presented with a text snippet and two event mentions highlighted. Figure~\ref{fig:mturk_example} shows two examples. To prune out event extraction errors from CAEVO, the annotators were first asked if the highlighted text denoted valid events. An event was deemed valid only if it describes an action that occurs in the world. \footnote{We left the decision for event validity up to annotators on their own. We asked annotators to consider an event with its context rather than the meaning of the word alone. This includes the negation of an event, which might imply a prevention relation.} If both triggers were deemed valid, then the annotators evaluated whether or not the candidate precondition event was an actual precondition for the target event. Specifically they check if the candidate event is necessary for the target event to happen.\footnote{We expected annotators to make decisions on the given CAEVO output, and they were not allowed to suggest a directional change. We limited the number of labeling options to keep the annotation instructions as straightforward as possible.}

We used a pilot task to refine the instructions and the examples to improve consistency amongst the annotators. For the main annotation task, we used \emph{four} crowd-workers to annotate each instance. For quality control, each HIT included control instances whose labels we knew {\em a priori}. We retained only those event pairs where a majority (i.e., at least three) of the annotators agreed on the label and use the majority label as the gold label for each instance.

\begin{table}[t!]
    \centering
    \begin{tabular}{l|c|c}
        & \nblue{Precond.} & \nblue{Non-Precond.}\\
        \hline\hline
        \#Evaluated & 200 & 200\\
        Errors & 13.5\% & 9\%\\
        - Event Validity & 1.5\% & 3.5\%\\
        - Relation & 12\% & 5.5\%\\ \hline
    \end{tabular}
    \vspace{-.1in}
    \caption{Expert review of PeKo annotations. "Event Validity" indicates annotation error on validity labels, "Relation" indicates errors on identifying the event-event relation.}
    \label{tab:annotation-quality}
    \vspace{-.2in}
\end{table}

\subsection{Dataset Quality and Analysis}

The resulting dataset, which we call \peko, contains more than 30K annotated relations (\textasciitilde10k preconditions, \textasciitilde20k not).

\vspace{.1in}
\noindent{\bf Annotation Quality}: The annotators had fair {\em inter-annotator agreement} with a Fleiss Kappa value $\kappa = 0.387$. We used 4 Turkers per event pair to ensure accuracy and filter out disagreements. To further measure the quality of the annotation, we randomly sub-sampled 400 instances from the annotated data and re-annotated them using four ``expert'' graduate students trained to recognize preconditions. A post-analysis of the expert and crowd annotations shows the annotation to be of high quality. Table~\ref{tab:annotation-quality} summarizes the quality statistics. Experts disagree with the crowd-sourced annotations in only $11.75\%$ of the cases, with a slightly higher disagreement for precondition instances at $13.5\%$. A small percentage of these disagreements are on determining when an event is valid.

We also analyzed the discarded instances that received conflicting votes. Only $10\%$ of these instances can be considered as preconditions and some of them are arguable based on their context. Here's an example:
\begin{quote}
Before he was \textbf{hired} in 2005, before his team upset Texas last season, he \textbf{educated} himself on the college culture.    
\end{quote}
According to the context with discourse cues, one can reasonably conclude that \textbf{educated} is necessary for the event \textbf{hired} to happen. However, one might also disagree based on the fact that the connection is not perfectly clear.
\\\vspace{-1mm}

\begin{figure*}[ht!]
    \centering{
    \begin{subfigure}{0.45\textwidth}
    \centering
    \includegraphics[width=\textwidth]{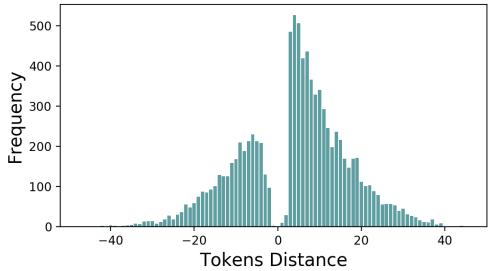}
    \label{fig:token_dist}
    \end{subfigure}
    \quad
    \begin{subfigure}{0.45\textwidth}
    \centering
    \includegraphics[width=\textwidth]{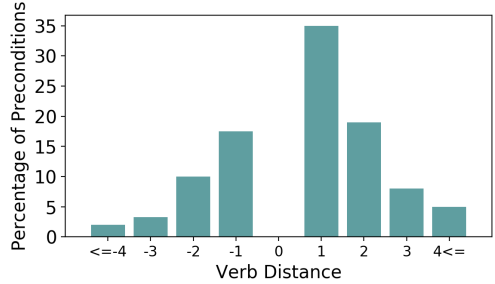}
    \label{fig:verb_dist}
    \end{subfigure}
    }
    \vspace{-.25in}
    \caption{
    Distribution of distances between preconditions and target events. Negative numbers correspond to cases where the target precedes the precondition, and positives for the other way around. The left plot shows the number of intervening tokens and the right shows percentages of verb distances between precondition and target events. }
    \label{fig:distance}
\end{figure*}

\noindent{\bf Text Position}: As with temporal and other event-event relations, one might ask if position in text is an indicator of a precondition relation. We thus tallied our annotations and identified how many intervening verbs occurred between the annotated event pairs, as well as how far apart they are in the document based on token distance.
Figure~\ref{fig:distance} shows these distributions. 
The negative numbers indicate distance when the precondition event occurs \emph{after} the target event. As the graphs show, the majority of preconditions occur first in the text, but a sizable amount are actually reversed with an evenly spread out distribution over distance. 

\begin{table}[ht!]
    \centering
    \begin{small}
    \begin{tabular}{p{0.18\textwidth}<{\centering}|p{0.18\textwidth}<{\centering}}
         \multicolumn{2}{c}{\nblue{Precondition Predicate $\rightarrow$ Target Predicate}} \\ \hline\hline
         pay $\rightarrow$ provide & try $\rightarrow$ get \\
         know $\rightarrow$ miss & ask $\rightarrow$ make \\
         use $\rightarrow$ provide  & love $\rightarrow$ miss\\
         go $\rightarrow$ provide & delay $\rightarrow$ mean  \\
         look $\rightarrow$ find & find $\rightarrow$ use \\
         take $\rightarrow$ get  & ask $\rightarrow$ take\\ 
         work $\rightarrow$ make & tell $\rightarrow$ take  \\
         use $\rightarrow$ find & know $\rightarrow$ get \\
         born $\rightarrow$ die  & agree $\rightarrow$ pay \\ 
         use $\rightarrow$ help &  touch $\rightarrow$ miss\\ 
         go $\rightarrow$ find &  get $\rightarrow$ help\\ 
         move $\rightarrow$ take &  lose $\rightarrow$ help\\ 
         leave $\rightarrow$ take \\ \hline
    \end{tabular}
    \end{small}
    \vspace{-.1in}
    \caption{The 25 most frequent predicate pairs in the annotated event pairs.}
    \label{tab:relations}
    \vspace{-.15in}
\end{table}


\eat{
\begin{table}[h]
    \centering
    \begin{tabular}{c|c}
         Distance & Percentage of Samples \\\hline\hline
         -3 & 0.02\\
         -2 & 0.05\\
         -1 & 5.74\\
         0 &  91.70 \\
         1 & 2.36\\
         2 & 0.12\\\hline
    \end{tabular}
    \caption{The distance between the sentences where the target and the source event are located}
    \label{tab:sent-dist}
\end{table}}

For further insight into the dataset, Table~\ref{tab:relations} lists the most frequent verbs that were annotated as precondition-target pairs. While there are a few pairs that can be readily interpreted without other context (e.g. everyone is born before they can die, and you must look before you can find), most other pairs require additional context from the text itself. 

\subsection{Comparison to Other Datasets}\label{sec:comparison}

Section \ref{sec:related} described how this new dataset differs from prior work. We now include Table \ref{tab:datasets} to further illustrate the size difference, showing an order of magnitude more precondition instances than prior corpora with specific precondition annotations.

We consider our precondition as a broader concept than that in the RED. We focus on \textit{necessary} events, which covers both precondition and causal relations in the RED dataset.

\begin{table}[ht!]
    \centering
    \begin{tabular}{l|ccc}
        \nblue{Dataset} & \nblue{\#Instances} & \nblue{\#Precond.}  \\\hline\hline
        RED (news/forums) &  4,969 & 1,055  \\
        CaTeRS (stories) &  2,715 & 488 \\
        StoryLine (news) & 12,423 & $< 5,519$*\\
        \textbf{\peko} (news)  & \textbf{28,948} & \textbf{10,806} \\ \hline 
    \end{tabular}
    \vspace{-.1in}
    \caption{Comparison of labeled corpora. The \emph{instances} are how many total labels, and \emph{precondition} is how many precondition-related instances. We included causation+precondition labels in the total counts if causation exists. *Event StoryLine mixes preconditions with many other relations, so the 5,519 is an upper bound. 
    }
    \vspace{-.2in}
    \label{tab:datasets}
\end{table}

\section{PeKo Tasks and Evaluation}

\begin{table*}[ht!]
    \parbox{.48\linewidth}{
    \centering
    \nblue{Precondition Identification}\\\vspace{.1in}
    \begin{tabular}{l|r|r|r}
        \nblue{Model} & \nblue{Precision} & \nblue{Recall} & \nblue{F1} \\ \hline \hline
        Random  & 37.34 & 50.00  & 42.75 \\
        GloVe-GRU & 56.25 & 73.38 & 63.68 \\
        BERT-feature & 59.80 & 81.15 & 68.84 \\
        XLNet & 66.69 & 77.10 & 71.52 \\ 
        BERT & 64.65 & 81.02 & 71.91 \\ \hline
    \end{tabular}
    \vspace{-.1in}
    \caption{Benchmarking performance of existing models on the precondition identification task. Simply fine-tuning large language models is not enough.}
    \label{tab:baseline-model}
    }
    \hfill
    \parbox{.48\linewidth}{
    \centering
    \nblue{Text Context Ablation}\\\vspace{.1in}
    \begin{tabular}{l|r|r|r}
        \nblue{Context} & \nblue{Precision} & \nblue{Recall} & \nblue{F1} \\ \hline \hline
        Event Trigger  & 54.06 & 75.68 & 63.07 \\
        Event Tuple     & 64.02 & 76.97 & 69.90 \\
        Event Tuple($\pm 1$)    & 63.84 & 78.95 & 70.59  \\
        Sentence        & 64.65 & 81.92 & 71.91 \\
        Sentence($\pm 1$) & 62.69 & 76.92 & 68.47\\
        Sentence($\pm 2$) & 61.65 & 77.33 & 68.60 \\\hline
    \end{tabular}
    \vspace{-.1in}
    \caption{Precondition identification results with varying levels of context using our BERT classifier.}
    \label{tab:context}
    }
    \vspace{-.1in}
\end{table*}

Having created the PeKo annotated corpus, we now propose two tasks that test for the ability to recognize and generate preconditions in textual contexts. Here we describe evaluations to benchmark the performance of current models on these tasks and to better understand the challenges involved. 
\subsection{PeKo Task 1: Precondition Identification}
Given a text snippet with a target and candidate event pair, the task is to classify if the candidate event is a precondition for the target in the context described by the text snippet. This is a standard sentence-level classification task. We evaluate two strong and widely-used large transformer-based language models -- fine-tuned BERT~\cite{devlin2019bert} and XLNet~\cite{yang2019xlnet} base models. For each model, we take the final representation of each event trigger, concatenate together, and then feed into a linear classification layer. We also evaluate a 1-layer GRU sequence model~\cite{cho2014properties} with GloVe embeddings~\cite{jeffreypennington2014glove} to calibrate against a much simpler baseline. See the Appendix for more details on parameters, layer sizes, and training time.

\vspace{.1in}
\noindent{\bf Precondition identification is a difficult task.}\\
Table~\ref{tab:baseline-model} shows the results. The GRU-based sequence model trained from scratch on \peko\ is better than a prior-based random baseline\footnote{Chooses a label at random from a binomial distribution of labels estimated from the training data} but still leaves a large room for improvement. BERT and XLNet both perform substantially better ($>71$ F1) than the GRU model (63.7 F1) but their F1 score of $71$ illustrates that this is a difficult task not readily solved by simply fine-tuning large LMs.

\vspace{.1in}
\noindent{\bf Precondition information is not readily available in BERT.}\\
One premise for our work is that distributional knowledge alone is insufficient to capture precondition relations. We conduct two sets of {\em inoculation-based probing experiments} (similar to \citet{liu-schwartz-smith:2019:NAACL}) to get at how the information in the pre-trained LM representations relate to preconditions. We use BERT in the fine-tuning and feature-extractor mode (the parameters for BERT are fixed and only those in the classification layer are updated) and measure performance with increasing amounts of data. 
If the performance peaks early with only small amounts of data then it tells us that most of the information necessary for recognizing preconditions is in a readily accessible form in the original LM representations. On the other hand, if performance keeps increasing then it suggests that \peko\ provides extra information. 
\begin{figure}[ht!]
    \centering
    \includegraphics[width=0.48\textwidth]{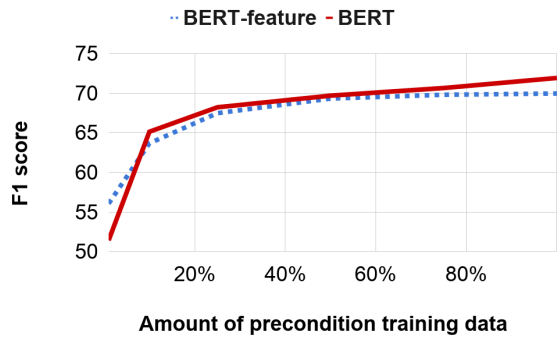}
    \vspace{-.3in}
    \caption{Inoculation: Performance of fine-tuning (solid) and feature extractor (dotted) modes of BERT with increasing amounts of \peko\ training data. Neither plateaus quickly suggesting that precondition knowledge is not readily accessible in BERT.}
    \label{fig:diff_data_size}
    \vspace{-.15in}
\end{figure}

Figure~\ref{fig:diff_data_size} shows that neither model plateaus quickly. BERT, as a feature-extractor (dashed line) plateaus around 50\% of the data. The fixed features from the LM pre-training BERT hits a performance ceiling. Whereas fine-tuning BERT, which fine-tunes the representation to the \peko\ task, provides continuous improvements for increasing amounts of data. These together suggest that a substantial amount of precondition knowledge is not easily adapted from the language modeling information captured in BERT but can be learned from \peko.


\vspace{.1in}
\noindent{\bf Role of Context.} Table \ref{tab:context} compares the performance of BERT when using different levels of context. Using event triggers alone achieves moderate performance. This suggests that the verb trigger does carry a lot of the precondition knowledge regardless of event arguments (e.g., canceling requires scheduling first, but in most cases it doesn't matter what is canceled). However, if we use event tuples\footnote{We used OpenIE\cite{Stanovsky2018SupervisedOI} to extract event tuples implemented in AllenNLP\cite{Gardner2017AllenNLP}}, which also captures the main entities of the event, then we see a significant improvement in performance (+6.9 points). In addition to the tuples of the event pair, adding tuple representations of neighboring events provides an additional gain (+1.5 points). 
Further inspection of the tuple-based representation shows that automatic tuple extraction sometimes introduces errors and misses critical context and other important discourse cues. The best results come from using the sentence(s) that contain the event pair in its entirety -- adding further sentences leads to worse performance. 

\begin{figure}[ht!]
    \centering
    \includegraphics[width=0.4\textwidth]{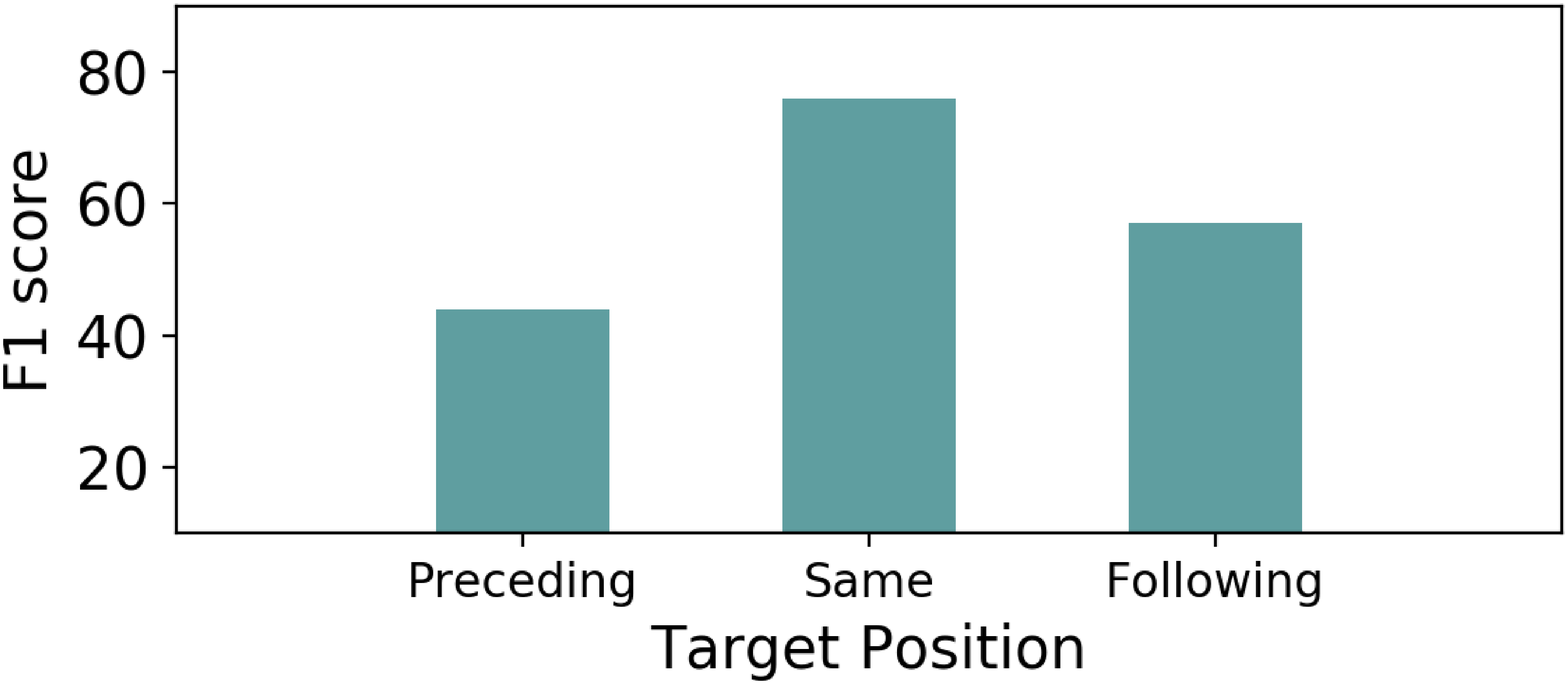} \\
    \includegraphics[width=0.4\textwidth]{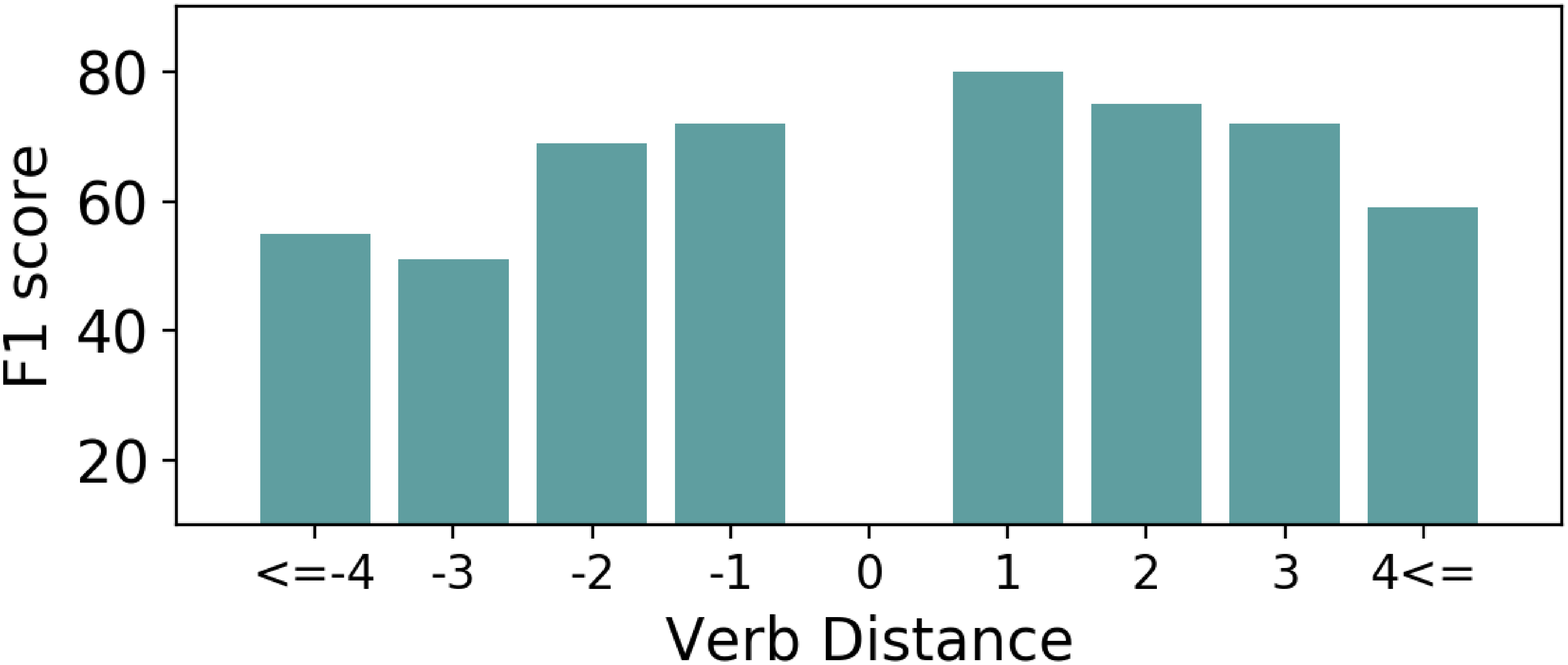}
    \caption{F1 scores across different contexts. Top: F1 when the target event precedes, is in the same, or follows the precondition's sentence. Lower: F1 for varying \# of intervening verbs between the event pair.
    }
    \label{fig:error-distributions}
    \vspace{-0.1in}
\end{figure}

\vspace{.1in}
\noindent{\bf When is it difficult to identify preconditions?}
The first plot in Figure~\ref{fig:error-distributions} 
shows that F1 score is highest where the target event is in the
same sentence as the precondition event, higher where the target event is in the sentence that follows the precondition event,  and lowest when the target event is in the previous sentence. 
A similar trend holds for different verb distances as well, as seen in the second plot. As the distance increases, the F1 score decreases in either direction. However, on the negative side, F1 scores are lower compared to the positive side showing the difficulty of the task when the target verb precedes the precondition.

\subsection{PeKo Task 2: Precondition Generation}
Here we introduce Precondition Generation as a more general challenge that a dataset like \peko\ now enables. Given a target event $t$, generate an event $p$ that is a precondition for $t$. We first show how to create instances for this task using the \peko\ dataset and then benchmark performance on evaluation instances drawn from both \peko\ and an out-of-domain dataset ATOMIC~\cite{sap2019atomic}.\\\vspace{-1mm}

\noindent{\bf Generation Training Task.} We created precondition generation training instances by transforming each \peko\ instance as follows. The \emph{input} is the entire snippet of a \peko\ instance (i.e, the entire text snippet with a pair of events where one is marked as a precondition of the other) but with the precondition portion of the snippet replaced by a \texttt{[BLANK]} slot. The \emph{output} for the generation instance is the entire sentence where the \texttt{[BLANK]} is to be filled in with text representing a precondition event. See Table \ref{tab:gen_examples} for examples. Note that because the precondition portion can occur anywhere (earlier or later) in the sentence, we do not frame this as a typical left-to-right language model completion task. Instead, the models have to generate the entire sentence in addition to filling in the \texttt{[BLANK]} slot with a plausible precondition. We use the text chunk spanned by the precondition trigger node in the constituency parse as the precondition portion.

We benchmark three variations of a large language model GPT-2~\cite{radford2019language} to show how much of precondition information can be generated directly from general language models and from temporal knowledge in comparison to learning from \peko: (i) {\bf LM-GPT-2} -- training instances created from a random collection of sentences to mimic fine-tuning GPT only for the format of this task but with no special constraint on the relation between the events in the instance. We randomly select sentences with a pair of events, and choose at random one event as target and the other as precondition and then create the generative training instances as described earlier. (ii) {\bf Temp-GPT-2} -- training on instances created from temporally BEFORE events, randomly sampled from the non-precondition portion of \peko\ dataset. (iii) {\bf \peko-GPT-2} -- training on generation instances created from the training portion of the \peko\ dataset. LM-GPT-2 trains on 18,000 instances (since it is not limited by \peko\ data), whereas Temp-GPT-2 and \peko-GPT-2 train on 6000 instances.\\\vspace{-1mm} 
 
\noindent{\bf Testing on \peko} For testing, we transform instances from the testing portion of \peko. Because precondition instances can sometimes contain strong linguistic and syntactic cues for preconditions, we create test instances only from the non-preconditions in \peko. This is a stronger test of models' abilities that mitigates some of the confounds of how the sentence is structured.\\\vspace{-1mm}

\noindent{\bf Testing on ATOMIC} We used the following heuristics to address the peculiarities of ATOMIC and improve compatibility with training. We filtered instances such that they are full sentences, with fully-specified arguments for events, and with single participant instances. We replace \texttt{Person} variable mentions with third-person pronouns.\\\vspace{-1mm}

\noindent{\bf Benchmarking Precondition Generation}. Table~\ref{tab:gen_manual} shows results of a manual evaluation of the generated preconditions\footnote{Automatic evaluation against reference preconditions is not informative since there can be multiple preconditions for any given event. We found that using BLEU for instance showed no difference between Temp-GPT2 and PeKo-GPT2 despite the huge difference in manual evaluation.}. Three of the authors of this paper evaluated 150 instances of generated text snippets from three systems. The snippets from the systems were randomly swapped during the blind evaluation.
Each output was first rated for sensibility on a scale of 0 to 3, where 3 means the output is perfectly sensible as English, and 0 means nonsensical.
The output, which contains the marked target and precondition event pairs, were then rated on a binary scale -- 1 if the precondition relation holds; 0 otherwise. The same annotation guidelines described in Section~\ref{sec:crowdsourcing} were taken to ignore invalid events, hypotheticals, and other noisy output.

\begin{table}[tb!]
    \centering
    \nblue{Precondition Generation}\\\vspace{.1in}
    \setlength\tabcolsep{3pt}
    \begin{tabular}{l|l|c|r}
         \nblue{Dataset} & \nblue{Model} & \nblue{Sense} & \nblue{Precond.}\\\hline\hline
          \multirow{3}{*}{\peko} & LM-GPT-2 & 1.69 & 12.00\% (12.87)\\
            & Temp-GPT-2 &  2.19 & 17.56\%  (17.21)\\
            & PeKo-GPT-2 & \bf{2.32} & \bf{35.81\% (37.96)} \\\hline
         \multirow{3}{*}{ATOMIC} & LM-GPT-2 & 2.20 & 10.40\% (10.78)\\
            & Temp-GPT-2 &  \bf{2.30} & 21.33\% (28.97) \\
            & PeKo-GPT-2 & 2.12 & \bf{39.33\% (50.43)} \\\hline   
    \end{tabular}
    \vspace{-.1in}
    \caption{Human evaluation of generation. Sense: Average sensibility rating on a 0-3 scale. Precond.: Percentage of instances with valid precondition outputs. Parenthetical numbers are precentages within instances with sensible score $\geq 2$. Bold face indicates best results.}
    \label{tab:gen_manual}
    \vspace{-.1in}
\end{table}

Results in Table~\ref{tab:gen_manual} shows that LM-GPT-2, the version that trains on random event pairs, struggles. It produces the least precondition outputs. Peko-GPT-2 generates plausible preconditions nearly \textbf{twice as often} as the Temp-GPT-2 baseline. These results illustrate the need for \peko\ as preconditions do not easily fall out from today's large LMs.
The trends also hold for the out-of-domain ATOMIC instances indicating generalization to everyday events in the ATOMIC dataset. On ATOMIC we see more preconditions than on the original \peko\ dataset. We hypothesize that this is in part because in the \peko\ test set, we created harder cases where the models have to generate preconditions to fit in text that originally contained a non-precondition event.

Table~\ref{tab:gen_examples} shows some examples that illustrate the differences between training on \peko\ and otherwise. As expected, the non-precondition trained model outputs events that temporally precede the target event but not necessarily preconditions.  \\\vspace{-1mm}

\noindent{\bf Error Analysis}. We evaluated the outputs for 50 instances from Peko-GPT-2 and found three main categories of failures: (i) Difficulty in handling input context (56\%) -- In some cases the input target event context is too limited, whereas in others the context is too complex with many intervening entities or a chained set of events after which the model is supposed to generate a precondition. Another set of cases have to do with the sentence structure of the context sets up for a hypothetical precondition event, or a reporting verb. (ii) Common Language Generation Errors (28\%) -- Cases like repetition or semantically implausible text and hallucinating new entities whose relation to the original context is not clear. (iii) Temporally related (16\%) -- Cases where the outputs are temporally and topically related but are not preconditions, indicating failures in generalizing precondition knowledge.

\begin{table}[tb!]
    \centering
    \begin{small}
    \begin{tabular}{|p{1.25cm}|p{5.6cm}|}
        \hline 
        \multicolumn{2}{|l|}{\nblue{\peko}} \\ \hline
        \textbf{INPUT} & [BLANK] that will \textit{enable} consumers to quickly download videos...\\
        Temp-G2 & The company has \textbf{\textit{said}} that it will be able to sell more phones in the coming months\\
        PeKo-G2 & The company also \textbf{\textit{agreed}} on a plan\\
        \hline
        \textbf{INPUT} & And finally a third will \textit{rebuild} homes in the historic older district. The foundations of a temporary market are in place, to house stores and stalls [BLANK]\\
        Temp-G2 & that have been \textbf{\textit{sold}} for decades. \\
        PeKo-G2 & that were \textbf{\textit{built}} in the 1880s. \\ \hline 
        \multicolumn{2}{l}{}\\ \hline
        \multicolumn{2}{|l|}{\nblue{ATOMIC}} \\ \hline
        \textbf{INPUT} & He \textit{is} in dire need of money [BLANK]\\
        Temp-G2 & he \textbf{\textit{said}} in an interview with The Daily.\\
        PeKo-G2 & because he has \textbf{\textit{lost}} his job. \\  \hline
        \textbf{INPUT} & She \textit{moves} to cambridge in 2013 [BLANK]\\
        Temp-G2 & when she \textbf{\textit{became}} the first woman to walk the halls of Congress. \\
        PeKo-G2 & she \textbf{\textit{took}} a job as a waitress at a local restaurant. \\ \hline
    \end{tabular}
    \end{small}
    \vspace{-.1in}
    \caption{Generation Examples on PeKo and ATOMIC test instances: {\bf INPUT} is the system input: text with the target event (italicised) and placeholder [BLANK]. Temp-G2 and PeKo-G2 are the generated outputs from the Temporal and \peko\ GPT-2 systems, with the precondition event in bold.}
    \label{tab:gen_examples}
    \vspace{-.1in}
\end{table}
Overall, these first results on \peko\ suggests that training on this new dataset enables a generative model to learn some common-sense precondition knowledge beyond basic language modeling cues. We see room for improvement both in terms of modeling as well as training approaches.

\eat{
\begin{table*}[ht!]
    \centering
    \begin{tabular}{r|r|r}
         Model & BLEU-1 (Input-Generated)  & BLEU-1 ([BLANK]-Generated) \\ \hline\hline
         Temp-GPT-2 & 0.624 & 0.107  \\
         PeKo-GPT-2 & 0.622 & 0.105  \\\hline
    \end{tabular}
    \caption{BLEU score of generation models. The numbers in the first column are the average score between input and generated text, and those in the second column are masked-out preconditions and generated text.\hk{This is possibly misguide readers, sine we can only compare the masked-out text with the entire generated text -- this will increase the denominator} We use Smoothing1~\cite{Chen2014ASC}}
    \label{tab:bleu}
\end{table*}
}

\section{Conclusions}
Knowing what conditions are necessary for an event to happen is critical for understanding and reasoning about events mentioned in text. In this work, we address the lack of a large scale resource for learning precondition knowledge about events. Our crowdsourcing methodology yielded more than 10,000 precondition event relations (and 20,000 negative examples) from news domain texts. We showed in both classification and generation that these relations are not readily accessible in distributional knowledge encoded by large language models, highlighting the challenges in learning common-sense knowledge from text.
We also proposed two new challenge tasks based on \peko\ and hope it helps drive further research into rich event understanding that touches a variety of areas from schema learning, information extraction, and even story generation. 

\section*{Acknowledgments}
 This material is based on research that is supported by the Air Force Research Laboratory (AFRL), DARPA, for the KAIROS program under agreement number FA8750-19-2-1003. The U.S. Government is authorized to reproduce and distribute reprints for Governmental purposes.

\bibliographystyle{acl_natbib}
\bibliography{tacl2018}
\newpage
\appendix
\section{Appendix}
\label{sec:appendix}

\subsection{Candidate Filtering for Crowdsourcing}
We discard event pairs that come from the same sentence when the candidate precondition is a causative verb or when the target is a reporting verb. This is because both cases are always true regardless of their context. Consider the following examples:

(A) He \textit{said} that his birth mother \textit{lived} nearby.

(B) The president \textit{made} his secretary \textit{create} copies of the report

As these examples show -- A is a reporting verb (`said') in the target position, and B is a causative (`made') as the candidate precondition -- the candidates in both cases are reliable preconditions independent of the context. For instance, in example in (B) if we use a new context ``not \textit{create} copies of the report'', the precondition relation would still hold. Since we aim to collect precondition knowledge that can be obtained at least partially from context, we excluded these reporting and causative precondition verb instances from our candidate pool.

\subsection{Experimental Details}

\subsubsection{Data Split}

We split our dataset into train/dev/test set with the ratio of 6:2:2. Since the number of instances in each class is imbalanced, we split the data separately based on the class and then randomly shuffle instances in each set together.

\subsubsection{Infrastructure}

All models are trained using NVIDIA Titan RTX (24GB of GDDR6 VRAM).

\subsubsection{Parameters}

\noindent \textbf{Identification Task}: All models for identification task are trained for 50 epochs with 16 of the batch size. A model is picked based on the performance (i.e., F1 score) on the dev set among 5 different random seeds. All other parameters are describe in Table \ref{tab:params}.

\noindent \textbf{Generation Task}: All three models use the same GPT-2 architecture, which has 163,047,936 trainable parameters. The epochs are set to 100 with 16 as the batch size. Models are picked based on loss on the dev set.

We use AdamW~\cite{loshchilov2018decoupled} for the optimizer in both tasks.

\begin{table}[h]
    \center
    \begin{tabular}{l|r|r}
         \nblue{Model} & \nblue{Hidden size} & \nblue{\#Parameters} \\\hline\hline
         GloVe-GRU & 512 & 9,675,154\\
         BERT-feature & 768 & 3,074\\
         XLNet & 768 & 116,721,410\\
         BERT &  768 & 108,313,346\\\hline
    \end{tabular}
    \caption{Parameters for the identification models. For GloVe-GRU model, we use GloVe embeddings with the size of 300.}
    \label{tab:params}
\end{table}

\subsubsection{Training Time}
Table \ref{tab:runtime} shows the training time for each model. The time is measured by the average elapsed time for each epoch excluding testing time on the dev set.

\begin{table}[h!]
    \center
    \begin{tabular}{c|l|r}
         \nblue{Task} & \nblue{Model} & \nblue{Time} \\\hline\hline
         \multirow{4}{*}{Identification} & GloVe-GRU & 25.29s \\
         & BERT-feature & 154.18s \\
         & XLNet & 204.15s \\
         & BERT & 235.85s \\\hline
         \multirow{3}{*}{Generation} & LM-GPT-2 & 574.99s\\
         & PeKo-GPT-2 & 126.83s \\
         & TEMP-GPT-2 & 130.20s \\ \hline
    \end{tabular}
    \caption{Average training time for each model on an epoch.}
    \label{tab:runtime}
\end{table}

\subsection{Testing on ATOMIC}
We following heuristics to address the peculiarities of the ATOMIC dataset and improve compatibility with training:
1) We remove instances that do not have a fully specified argument for the event (referred to as placeholders in their paper~\citep{sap2019atomic}). 2) We only use `simple' instances that mention a single participant because the context often contains enough information to fully understand the target event. 3) We only use instances that are complete sentences and not fragments. 4) To make the inputs more natural, we replace the \texttt{Person} variable mentions with a third-person pronoun and added markers to the main verb and the placeholder \texttt{[BLANK]} at the end:

\emph{``PersonX is in dire need of money''} to \emph{``He <target> is </target> in dire need of money [BLANK]''}

\end{document}